\documentclass{article}
\usepackage{arxiv}

\usepackage[utf8]{inputenc}
\usepackage[T1]{fontenc}
\usepackage[numbers]{natbib}
\usepackage[hidelinks]{hyperref}
\usepackage{url}
\usepackage{booktabs}
\usepackage{enumitem}
\usepackage{amsfonts}
\usepackage{amsmath}
\usepackage{amssymb}
\usepackage{microtype}
\usepackage{graphicx}
\usepackage{float}
\usepackage{tikz}
\usetikzlibrary{svg.path}

\renewcommand{\headeright}{}
\renewcommand{\undertitle}{}

\definecolor{orcidlogocol}{HTML}{A6CE39}
\tikzset{orcidlogo/.pic={
  \fill[orcidlogocol] svg{M256,128c0,70.7-57.3,128-128,128C57.3,256,0,198.7,0,128C0,57.3,57.3,0,128,0C198.7,0,256,57.3,256,128z};
  \fill[white] svg{M86.3,186.2H70.9V79.1h15.4v48.4V186.2z}
    svg{M108.9,79.1h41.6c39.6,0,57,28.3,57,53.6c0,27.5-21.5,53.6-56.8,53.6h-41.8V79.1z M124.3,172.4h24.5c34.9,0,42.9-26.5,42.9-39.7c0-21.5-13.7-39.7-43.7-39.7h-23.7V172.4z};
  \fill[white] svg{M88.7,56.8c0,5.5-4.5,10.1-10.1,10.1c-5.6,0-10.1-4.6-10.1-10.1c0-5.6,4.5-10.1,10.1-10.1C84.2,46.7,88.7,51.3,88.7,56.8z};
}}
\newcommand{\orcidlink}[1]{\href{https://orcid.org/#1}{\,\raisebox{-0.15ex}{\resizebox{!}{0.85em}{\begin{tikzpicture}[yscale=-1]\pic{orcidlogo};\end{tikzpicture}}}}}

\title{A JoLT for the KV Cache:\\ Near-Lossless KV Cache Compression via Joint Rank-bit Allocation}

\author{%
  Rahul Krishnan\,\orcidlink{0009-0000-4542-7989} \\
  Universit\"at Trier
  \And
  Volker Schulz\,\orcidlink{0000-0001-7665-130X} \\
  Fachbereich IV, Mathematik, Universit\"at Trier
}

\begin{document}
\maketitle


\begin{abstract}
The key-value (KV) cache is the dominant memory bottleneck in long-context language model inference.
Existing compression methods apply low-rank factorization or quantization independently, without jointly allocating rank and precision under a shared storage budget.
We introduce JoLT, a training-free compressor that treats grouped prefill caches as fourth-order tensors and applies partial Tucker decomposition along the token and feature modes, the two axes that carry low-rank structure, while leaving the head and layer modes intact.
A rotated low-bit quantizer captures the truncation residual, and a single Lagrangian dual allocates per-group Tucker ranks and residual bit-widths under a global byte constraint.
FlashJoLT replaces the exact token-mode SVD with a randomized approximation that matches JoLT within the free zone at a fraction of the compression cost, and a fused Triton decode kernel evaluates attention directly over the stored factors without materializing dense KV tensors.
Across five models from four architecture families, covering multi-head attention, grouped-query attention, and mixture-of-experts architecture, JoLT achieves $2$--$3\times$ compression with less than $0.2\%$ perplexity degradation, without retraining.
On RULER at $64$K context with LLaMA-3.1-8B, retrieval accuracy remains near-lossless through $3\times$ and declines by only $0.90$ and $2.40$~pp at $4\times$ and $5\times$, respectively.
JoLT demonstrates that tensor-aware low-rank decomposition and quantized residuals, unified under a single storage budget, achieve near-lossless KV-cache compression across diverse model architectures without retraining.

\end{abstract}


\section{Introduction}
\label{sec:intro}

Autoregressive decoding reuses the keys and values of every preceding token at every layer, so inference systems store them in a key-value (KV) cache~\citep{vaswani2017attention}.
The cache grows linearly with batch size, context length, layer count, and the number and dimension of KV heads, and at large batch sizes and long contexts it outgrows the model weights themselves~\citep{pope2023efficiently}.
Reducing it is the most direct way to serve longer contexts and larger batches on a given device~\citep{kwon2023pagedattention}.

Existing compression methods shrink the cache through low-rank projection or quantization.
Low-rank methods such as xKV~\citep{chang2025xkv} factor a single two-dimensional unfolding of the cache, capturing correlation across layers but not across other axes.
Quantization methods such as KIVI~\citep{liu2024kivi} and TurboQuant~\citep{turboquant2025} keep every entry of the full tensor at lower precision and leave its low-rank structure unused.
Hybrid methods combine the two but fix the split by hand: GEAR~\citep{kang2024gear} corrects a quantized cache with a low-rank term of preset rank, and Palu~\citep{chang2024palu} and STAR-KV~\citep{bhatnagar2026starkv} quantize their low-rank latents at a preset bit-width after the ranks are chosen.
None of them decides how much of a fixed byte budget to spend on rank and how much on precision.

\citet{krishnan2026comparative} treat grouped layer caches as fourth-order tensors over layers, heads, tokens, and features, and measure the singular-value spectrum of each mode.
The head and layer modes are index-like, with nearly flat spectra (a dynamic range of $1.4$ on the head mode of Mistral-7B), so truncating them incurs a large error that no budget can recover.
The token and feature modes carry the low-rank structure, and values are about $2.4\times$ harder to compress than keys because their feature spectrum is flat.
These findings determine both where to truncate and how to split the budget.

\begin{figure}[t]
    \centering
    \includegraphics[width=0.9\linewidth]{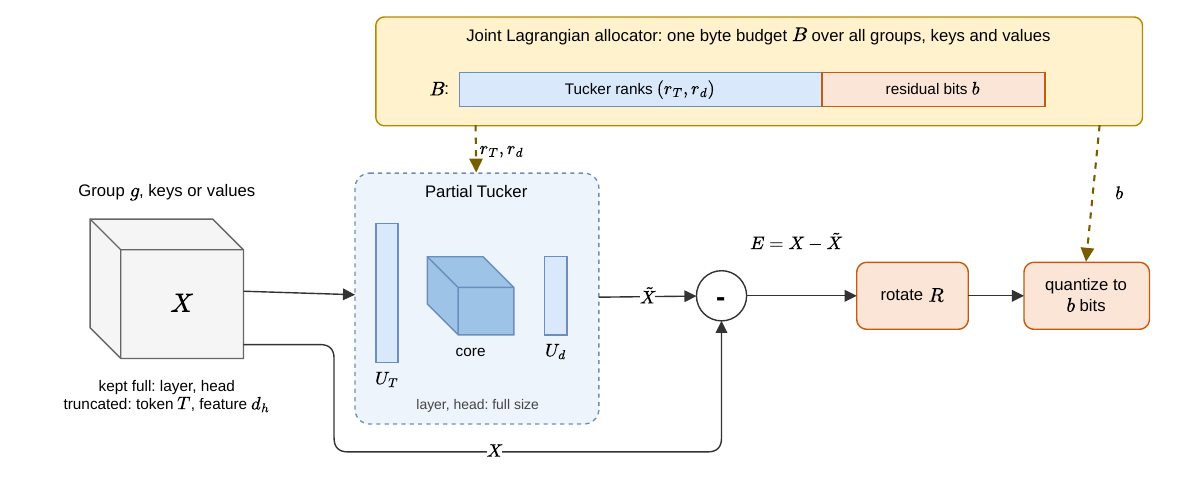}
    \caption{Overview of JoLT. Each cell (one layer group's keys or values) is compressed in two stages, with a single Lagrangian allocator splitting the byte budget $B$ across Tucker ranks and residual bit-widths for all cells jointly.}
    \label{fig:overview}
\end{figure}

We introduce \textbf{JoLT} (Joint Lagrangian Tucker), a training-free compressor built on this design (Figure~\ref{fig:overview}).
JoLT groups consecutive layers whose token subspaces align, so that one pair of token and feature bases serves the whole group, and applies a partial Tucker decomposition that truncates only the token and feature modes.
The truncation discards a spectral tail that is large especially for values, so JoLT stores it as a low-bit residual after a random orthogonal rotation spreads channel outliers across coordinates.
A single Lagrangian dual then allocates Tucker ranks and residual bit-widths to every group, and to keys and values separately, under one global byte budget.
Each byte therefore goes to the rank or the bit that removes the most reconstruction error.

Two further components make JoLT practical.
FlashJoLT replaces the exact token-mode SVD, which dominates compression time, with a randomized SVD~\citep{halko2011randomized} that computes only the leading singular values.
It estimates the energy of the remaining ones from the tensor norm so that the allocator's rank-bit decisions stay correct.
Decompressing the cache to dense tensors at decode time would erase the memory savings, so a fused Triton~\citep{tillet2019triton} kernel evaluates attention directly over the stored factors and packed residuals.

Across five models from four families, spanning multi-head~\citep{vaswani2017attention} and grouped-query attention~\citep{ainslie2023gqa} as well as a mixture-of-experts architecture~\citep{muennighoff2024olmoe}, JoLT compresses the cache $2$--$3\times$ with less than $0.2\%$ perplexity drift and no retraining.
We call this band the free zone.
The free zone carries over to downstream tasks: on RULER~\citep{hsieh2024ruler} at $64$K context with LLaMA-3.1-8B, retrieval stays near-lossless through $3\times$ and drops by only $0.90$ and $2.40$~pp at $4\times$ and $5\times$ (Sections~\ref{sec:results} and~\ref{sec:downstream}).

Our contributions are:
\begin{enumerate}[leftmargin=*,itemsep=3pt,parsep=0pt,topsep=4pt]
  \item \textbf{Joint rank-precision allocation.} A compression formulation that combines partial Tucker truncation of the token and feature modes with a rotated low-bit residual, and allocates ranks and bit-widths jointly under one byte budget (Section~\ref{sec:method}).
  \item \textbf{FlashJoLT.} A randomized token-mode decomposition with tail-mass accounting method that keeps the allocation correct and matches JoLT in the free zone at a fraction of the compression cost (Section~\ref{sec:fast}).
  \item \textbf{A cross-architecture free zone.} Evidence on five models that $2$--$3\times$ compression is near-lossless without retraining and holds on downstream reasoning and long-context retrieval (Sections~\ref{sec:results} and~\ref{sec:downstream}).
  \item \textbf{Fused decode kernel.} A fused Triton kernel evaluates attention over the compressed representation without dense reconstruction, reducing peak KV decode memory by $1.43$ to $4.73\times$ across $2$--$8\times$ targets (Section~\ref{sec:efficiency}).
\end{enumerate}


\section{Related Work}
\label{sec:related}

\paragraph{KV-cache quantization.}
Quantization compresses the cache by lowering the bit-width of every stored entry.
KIVI applies tuning-free asymmetric $2$-bit quantization with per-channel keys and per-token values~\citep{liu2024kivi}.
TurboQuant composes a random rotation with an MSE-optimal scalar quantizer and a $1$-bit residual projection to approach the rate-distortion bound~\citep{turboquant2025}.
KVQuant quantizes keys before the rotary embedding and handles per-channel outliers~\citep{hooper2024kvquant}, sharing JoLT's pre-RoPE motivation but fixing the bit-width in advance with no rank component.
Each quantizer operates at one fixed ratio set by its bit-width, with no way to target intermediate values.
JoLT covers the $2$--$10\times$ range continuously by jointly optimizing rank and precision at any target byte budget. The two approaches are complementary: the model weights and the Tucker factors themselves can be quantized for additional compression beyond what either method achieves alone.

\paragraph{Low-rank and structural KV compression.}
Palu factorizes per-head-group projection weights offline and caches the low-rank latents, optionally quantized after a fused Hadamard rotation~\citep{chang2024palu}.
xKV projects grouped cross-layer feature blocks onto a shared singular-vector basis~\citep{chang2025xkv}.
KQ-SVD decomposes the attention matrix $QK^\top$ rather than the cache itself~\citep{lesens2025kqsvd}.
ReCalKV combines head-reordered SVD for keys with calibrated value projections~\citep{yan2025recalkv}.
STAR-KV picks per-head key ranks by learned soft thresholding and quantizes the latents at a fixed mixed-precision split~\citep{bhatnagar2026starkv}.
All of these are matrix methods: each factors a single two-dimensional slice of the cache and never sees the full multi-linear structure across heads, tokens, features, and layers.
Palu and STAR-KV add quantization to their latents, but both fix the bit-width by hand after the ranks are set.
GEAR~\citep{kang2024gear} takes the opposite approach, pairing a quantized base with a low-rank correction and a sparse outlier term, but fixes its rank, sparsity ratio, and bit-width uniformly across layers.
In all cases, rank and precision are set independently rather than traded off under a shared budget.
Token-eviction methods are orthogonal in mechanism and can in principle be combined with JoLT~\citep{zhang2023h2o,xiao2024streamingllm,li2024snapkv}.

\paragraph{Tensor decompositions and rate allocation.}
JoLT computes its partial Tucker decomposition via the sequentially truncated higher-order SVD (ST-HOSVD)~\citep{vannieuwenhoven2012sthosvd}, a sequential variant of the HOSVD~\citep{delathauwer2000hosvd} applied to the Tucker format~\citep{tucker1966some}.
CP~\citep{kolda2009tensor}, tensor train~\citep{oseledets2011tt}, and t-SVD~\citep{kilmer2011tsvd} are alternative tensor formats.
JoLT uses Tucker because it can leave the index-like head and layer modes untouched while truncating only the compressible ones.
The rotated low-bit residual draws on incoherence processing~\citep{chee2023quip}, the principle behind rotation-based weight quantization~\citep{ashkboos2024quarot,liu2025spinquant}, applied here to the truncation residual of a tensor factorization rather than to weights.
The Lagrangian allocation that ties the two stages together is a standard relaxation of a constrained resource-allocation problem, but has not previously been applied jointly to per-mode Tucker ranks and residual bit-widths for KV-cache compression.

\section{Method: Joint Lagrangian Tucker (JoLT)}
\label{sec:method}

The spectral structure described in Section~\ref{sec:intro} determines JoLT's design: truncate only the token and feature modes, leave the head and layer modes at full rank, and give keys and values separate budgets so that the harder value tensor draws more bytes.
JoLT compresses the KV cache produced during prefill in two stages, a partial Tucker decomposition and a rotated low-bit residual, and allocates storage to each stage with a single Lagrangian optimizer (Figure~\ref{fig:overview}).

We partition the $L$ layers into $G$ contiguous groups by clustering layers whose token-mode singular vectors are well-aligned, so that a single set of factor bases can serve the group without cross-layer interference.
For each layer, we compute the top eight token-mode singular vectors of its key cache and score consecutive layers by the mean squared cosine between their bases.
A new group starts whenever the score falls below the median score across the network.%
\footnote{The size bounds take priority: a new group opens only once the current one has at least two layers, a group is closed at eight layers regardless of the score, and a trailing group of one layer is merged into its predecessor, so the last group can reach nine layers.}
The grouping is computed from the cache being compressed, with no separate calibration data.
On Mistral-7B-v0.3 at $T{=}1024$, the $32$ layers form $11$ groups.

Keys are decomposed before the rotary position embedding~\citep{su2021roformer} because post-RoPE keys show $48$ to $62\%$ higher reconstruction error at matched storage~\citep{krishnan2026comparative}.
The rotation is applied to the reconstructed keys at decode time.

Within each group $g$, we treat keys and values as separate compression targets $t \in \{\mathrm{K}, \mathrm{V}\}$.
Each group's key or value cache forms a fourth-order tensor $\mathcal{X}_g \in \mathbb{R}^{|g| \times n_h \times T \times d_h}$, where $|g|$ is the number of layers in the group, $n_h$ is the number of KV heads, $T$ is the number of cached prefill tokens, and $d_h$ is the per-head feature dimension.
The partial Tucker decomposition truncates the token and feature modes to ranks $r_T$ and $r_d$ while keeping the layer and head modes at full rank with identity factors.
We call each pair $(g, t)$ a \emph{cell} and compress every cell with the three stages below.

\paragraph{Partial Tucker backbone.}
The partial Tucker approximation $\tilde{X}(r_T, r_d)$ computes the mode-$k$ singular vectors of the token and feature unfoldings via the sequentially truncated higher-order SVD (ST-HOSVD)~\citep{vannieuwenhoven2012sthosvd} and truncates them to ranks $r_T$ and $r_d$, while keeping identity factors on the layer and head modes.
The token mode is truncated first, and the feature-mode factor is then computed on the projected tensor, so the feature truncation sees only the energy that survived the token step.
The resulting core has shape $|g| \times n_h \times r_T \times r_d$, and two factor matrices $U_T \in \mathbb{R}^{T \times r_T}$ and $U_d \in \mathbb{R}^{d_h \times r_d}$ store the token-mode and feature-mode bases shared across every head and every layer in the group.
With identity factors on the layer and head modes, a single ST-HOSVD pass directly constructs the partial Tucker approximation without iterative refinement.

The ST-HOSVD approximation is quasi-optimal in Frobenius norm: its error is at most $\sqrt{2}$ times the minimum over all approximations with token rank at most $r_T$ and feature rank at most $r_d$.
This follows from the ST-HOSVD error bound~\citep[Thm.~6.5]{vannieuwenhoven2012sthosvd} and the Eckart--Young theorem on each of the two truncated unfoldings.
Riemannian optimization on the fixed-rank Tucker manifold~\citep{heidel2018riemannian} could tighten this bound, but we keep the single pass for speed and let the residual stage capture the full truncation error.

\paragraph{Rotated low-bit residual.}
The truncation discards a spectral tail that is large for values, so moderate Tucker ranks alone cannot reach near-lossless fidelity.
Rather than spending additional rank on diminishing returns, JoLT stores the truncation residual $E = \mathcal{X}_g - \tilde{X}(r_T, r_d)$ at low bit-width.

We apply a random orthogonal rotation along the feature axis to spread residual energy evenly across channels, so that no single outlier channel dominates the quantizer's dynamic range.
The rotation matrix is the $d_h \times d_h$ factor from the QR decomposition of a seeded Gaussian, shared by every cell and regenerated from its seed at decode, so it costs no storage.
We then quantize the rotated entries uniformly at $b \in \{0, 2, 4, 8\}$ bits per element with one fp32 scale per row, where $b{=}0$ means no residual is stored.
Decoding inverts the rotation and adds the dequantized residual to the low-rank reconstruction.

\paragraph{Joint Lagrangian allocation.}
The backbone and the residual compete for the same bytes, and the right split depends on the spectrum of each cell.
A fixed policy that always applies four residual bits, or never stores a residual, cannot adapt to this per-cell variation.
JoLT therefore allocates ranks and residual bits jointly under a single byte budget.

Consider a single cell $(g, t)$ and let $c$ denote the bytes per stored scalar ($c{=}2$ for bf16).
A choice of token rank $r_T$, feature rank $r_d$, and residual bit-width $b$ costs
\begin{equation}
  s_{g,t}(r_T, r_d, b) \;=\; \underbrace{\big( |g|\, n_h\, r_T r_d + T r_T + d_h r_d \big)\, c}_{\text{Tucker core and factors}} \;+\; \underbrace{[\,b > 0\,]\,\frac{b + 32/d_h}{8}\, |g|\, n_h\, T\, d_h}_{\text{packed rotated residual}}
  \label{eq:bytes}
\end{equation}
bytes.
The two identity factors on the layer and head modes are negligible and omitted.
The residual term accounts for one fp32 scale per residual row, giving an effective bit-width of $b + 32/d_h$.
At $d_h{=}128$ and $b{=}4$ this is $4.25$ bits per element, and at $b{=}0$ no residual and no scales are stored, so the term vanishes.

The reconstruction error is modeled as
\begin{equation}
  e_{g,t}(r_T, r_d, b) \;\approx\; \varepsilon^2(b)\,\cdot\,\tau_{g,t}(r_T, r_d) .
  \label{eq:err}
\end{equation}
Here $\tau_{g,t}(r_T, r_d)$ is the relative Frobenius mass that the partial Tucker truncation discards at ranks $(r_T, r_d)$, computed as the sum over the two truncated modes of the discarded mode-$k$ singular-value mass divided by $\|\mathcal{X}_g\|_F^2$.
The factor $\varepsilon^2(b)$ is the fraction of that discarded mass the rotated $b$-bit quantizer fails to recover, calibrated once on a Gaussian round-trip with $\varepsilon^2(0){=}1$.
At $d_h{=}128$ the calibration gives $\varepsilon^2(2)\approx0.59$, $\varepsilon^2(4)\approx0.014$, and $\varepsilon^2(8)\approx4\times10^{-5}$.
This multiplicative factorization is the key modeling choice: rank reduces the truncation tail $\tau$, and bits reduce the residual factor $\varepsilon^2(b)$, making the two stages commensurable on a single error scale.

\paragraph{Solving the allocation.}
Given a total byte budget $B$ for the compressed prefill cache, JoLT minimizes the summed error subject to the summed cost:
\begin{equation}
  \min_{\{(r_T, r_d, b)_{g,t}\}} \; \sum_{g,t} e_{g,t}(r_T, r_d, b)
  \quad \text{s.t.} \quad \sum_{g,t} s_{g,t}(r_T, r_d, b) \le B .
  \label{eq:problem}
\end{equation}
We solve this through its Lagrangian relaxation,
\begin{equation}
  \mathcal{L}(\lambda) \;=\; \sum_{g,t} \Big[\, e_{g,t}(r_T, r_d, b) + \lambda\, s_{g,t}(r_T, r_d, b) \,\Big],
  \label{eq:lagrangian}
\end{equation}
which separates across cells.
Because $r_T$, $r_d$, and $b$ lie on finite grids, Equation~\eqref{eq:problem} is a discrete separable allocation problem and requires no differentiability.

For fixed $\lambda$, each cell independently minimizes $e + \lambda s$.
We enumerate $b \in \{0,2,4,8\}$ and, for each $b$, greedily increase ranks starting from $r_T = r_d = 1$.
At each step the solver compares the marginal error reduction per byte for incrementing $r_T$ by one against incrementing $r_d$ by one, and takes whichever gives the steeper descent.
A mode's rank increases by one whenever $\varepsilon^2(b)\,\sigma_{k,r_k+1}^2/\|\mathcal{X}_g\|_F^2 > \lambda\,\Delta s$, where $\sigma_{k,\cdot}$ are the singular values of the unprojected mode-$k$ unfolding and $\Delta s$ is the byte increment from Equation~\eqref{eq:bytes}.
The search stops when a full sweep over both modes makes no change, and the cell keeps the $b$ with the smallest $e + \lambda s$.
This rule is exact for a single truncated mode but is a coordinate-wise heuristic with two modes because the core cost contains the bilinear term $|g|\, n_h\, r_T r_d$.
With exact per-cell minimization, the multiplier theorem of \citet{everett1963generalized} guarantees optimality for the budget actually attained, following the standard approach in discrete bit allocation~\citep{shoham1988efficient}.
Since total cost is non-increasing in $\lambda$, we find the multiplier by geometric bisection, stopping within $0.5\%$ of the target budget or after $50$ iterations.
The multiplier $\lambda$ acts as a global shadow price on memory, jointly balancing rank and residual precision across all cells.

\paragraph{Surrogate limitations.}
The surrogate of Equation~\eqref{eq:err} charges every unit of discarded Frobenius mass equally, whether it remains as structured truncation error or as quantization noise, and it does not model the asymmetric effect of key errors through the softmax.
In the free zone this approximation is accurate, but at higher compression on LLaMA-2 it makes extra key rank appear cheaper than key residual bits of the same byte cost.
Section~\ref{sec:fast} introduces a token-rank cap that guards against this failure mode, and Section~\ref{sec:ablations} quantifies the effect.

\subsection{Fast variant: FlashJoLT}
\label{sec:fast}

The exact token-mode SVD dominates compression time.
The token-mode unfolding of $\mathcal{X}_g \in \mathbb{R}^{|g| \times n_h \times T \times d_h}$ has shape $T \times (|g|\, n_h\, d_h)$, and its economy SVD scales as $O\!\bigl(T\,(|g|\, n_h\, d_h)^2\bigr)$ when $T$ exceeds $|g|\, n_h\, d_h$, which is the typical regime at long contexts.
FlashJoLT replaces this SVD with a randomized approximation~\citep{halko2011randomized} that computes only the leading $q_\mathrm{cap}$ token-mode singular vectors.
The feature-mode SVD is kept exact because $d_h$ is small ($128$), and the rest of the pipeline, including the Lagrangian allocator, is unchanged.

The allocator requires the full spectrum to price rank against residual bits.
Since the randomized SVD computes only the top $q_\mathrm{cap}$ singular values, FlashJoLT estimates the remaining energy as $\|\mathcal{X}_g\|_F^2$ minus the energy of the computed singular values and appends it as a single synthetic tail entry.
The allocator therefore sees the correct total energy and makes valid rank-bit decisions on the truncated spectrum.

The cap $q_\mathrm{cap}$ controls how many singular vectors are computed.
With $C$ the target compression ratio, the policy is
\begin{equation}
  q_{\mathrm{cap}} \;=\; \min\!\bigl(\max(q_{\min}(C),\, \lceil T/32 \rceil),\, 512\bigr), \qquad q_{\min}(C) = \begin{cases} 32 & C \le 4 \\ 64 & C > 4, \end{cases}
  \label{eq:qcap}
\end{equation}
so the cap starts at $q_\mathrm{min}(C)$ for short contexts, grows linearly with $T$, and saturates at $512$.
The cap also serves as a guard against the surrogate limitation described in Section~\ref{sec:method}: by bounding the token rank, it prevents the allocator from over-spending on key rank at the expense of key residual bits, which matters on LLaMA-2 past the free zone (Appendix~\ref{app:fast}).
Calibration details and the LLaMA-2 analysis are in Appendix~\ref{app:fast}.

Quality is essentially unchanged in the free zone.
On Mistral at $T{=}1024$, the reconstruction-error gap between FlashJoLT and JoLT is $|\Delta_K| \le 0.0093$ and $|\Delta_V| \le 0.0011$ across $2$--$3\times$ (Appendix~\ref{app:fast}).
Timing both compressors over the full compression pipeline, including layer grouping, gives end-to-end speedups of $1.04\times$ to $3.97\times$ depending on cache length and target ratio (Appendix~\ref{app:fast}).

\section{Experiments}
\label{sec:results}

\subsection{Setup}

We evaluate JoLT on five models spanning four architecture families:
Mistral-7B-v0.3~\citep{jiang2023mistral} (GQA, $n_\mathrm{kv}{=}8$, $32$ layers),
LLaMA-2-13B~\citep{touvron2023llama2} (MHA, $n_\mathrm{kv}{=}40$, $40$ layers),
LLaMA-3.1-8B~\citep{grattafiori2024llama3} (GQA, $n_\mathrm{kv}{=}8$, $32$ layers),
Qwen2.5-14B~\citep{qwen2025qwen25} (GQA, $n_\mathrm{kv}{=}8$, $48$ layers),
and OLMoE-1B-7B-0924~\citep{muennighoff2024olmoe} (MoE with MHA, $n_\mathrm{kv}{=}16$, $16$ layers).
Mistral and LLaMA-2 are the primary evaluation pair and appear in every experiment.

All models run in bf16 on A100 GPUs (see the Reproducibility Statement) with float32 decomposition numerics.
At context length $n$, the prompt is split into two halves: the first $n/2$ tokens serve as the prefill cache that JoLT compresses ($T = n/2$), and the second half is scored against the compressed cache to produce the perplexity.
Perplexity is computed on a deterministic WikiText-2~\citep{merity2017wikitext} and C4~\citep{raffel2020c4} mixture and reported as the mean over seeds $\{0,1,2\}$ unless otherwise noted.
Across-seed standard deviation is at most $0.009$~PPL on every free-zone cell.

Unless stated otherwise, results use JoLT (exact SVD).
FlashJoLT is used for LLaMA-3.1 and Qwen in the perplexity tables and for all RULER experiments.

\subsection{Perplexity: the near-lossless free zone}
\label{sec:ppl}

We call the $2$--$3\times$ compression band the \emph{near-lossless free zone}: the regime where perplexity drift stays below $0.2\%$ on every model tested.

\begin{figure}[t]
  \centering
  \includegraphics[width=1\linewidth]{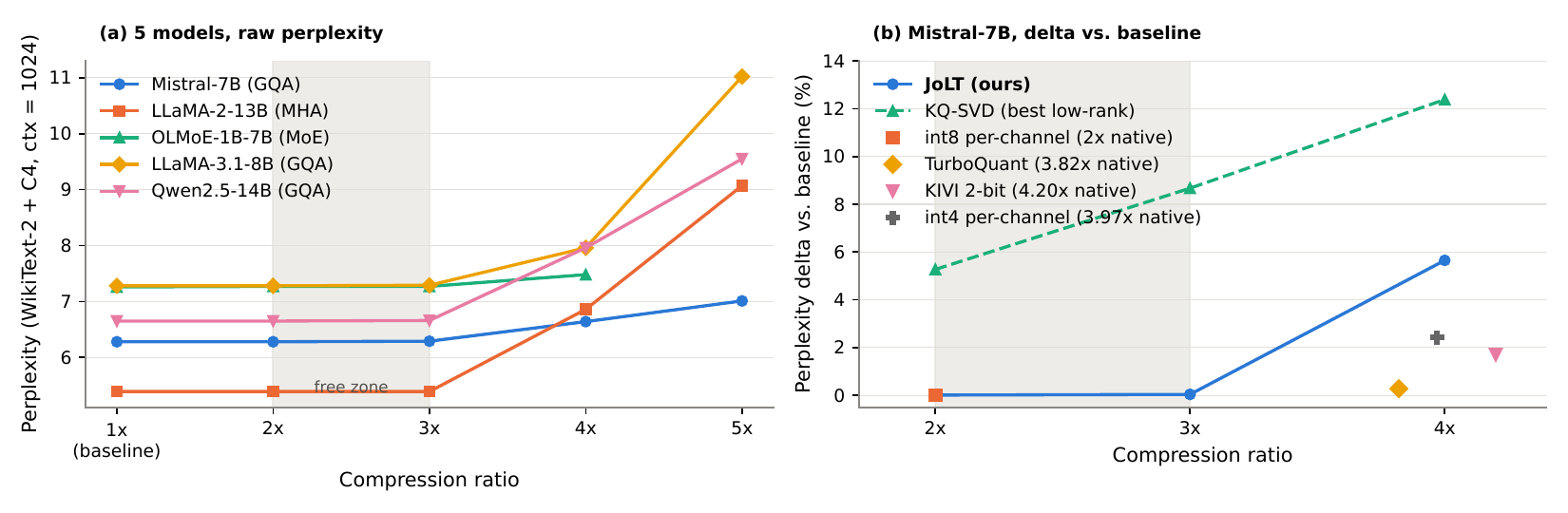}
  \caption{(a) Perplexity versus compression ratio at $n{=}1024$ on all five models (Table~\ref{tab:freezone}). The $2$--$3\times$ free zone is shaded. (b) Perplexity change on Mistral-7B for JoLT and baselines on the $150$-chunk cross-method grid (Table~\ref{tab:baselines}). Quantizers are shown at their native ratio.}
  \label{fig:freezone}
\end{figure}

Table~\ref{tab:freezone} reports perplexity at $n{=}1024$ ($T{=}512$) on all five models.
Every free-zone cell stays below the $0.2\%$ threshold, with the largest drift at $+0.17\%$ (Qwen at $3\times$).
The free zone holds at context lengths $512$ and $2048$ as well (Appendix~\ref{app:ppl}).
Beyond the free zone the models diverge: Mistral and OLMoE degrade gradually at $4\times$ ($+5.59\%$ and $+3.00\%$), while LLaMA-2 jumps sharply to $+27.28\%$ (Figure~\ref{fig:freezone}a).
The degradation pattern is model-specific, not attention-type-specific, and a detailed breakdown appears in Appendix~\ref{app:ppl}.
\paragraph{Comparison with baselines.}
Table~\ref{tab:baselines} compares JoLT against six low-rank baselines, three fixed-rate quantizers, and two $8$-bit baselines on Mistral at $n{=}1024$, evaluated on a separate $150$-chunk grid (LLaMA-2 results in Appendix~\ref{app:recon}, Table~\ref{tab:ppl_baselines_llama}).
At $2\times$, JoLT drifts $+0.01\%$ while the best low-rank baseline (KQ-SVD) drifts $+5.27\%$.
Fixed-rate $8$-bit quantization (int8 per-channel and fp8 e4m3) is also near-lossless at $2\times$, but has no operating point above $2\times$.
Near $4\times$, the dedicated quantizers at their native ratios outperform JoLT: TurboQuant at $+0.28\%$ and KIVI at $+1.70\%$ against JoLT's $+5.65\%$.
JoLT's advantage over every evaluated baseline therefore lies in the band above $2\times$ and below roughly $4\times$, which is exactly the free zone (Figure~\ref{fig:freezone}b).

\begin{table}[t]
  \centering
  \caption{Perplexity at context length $n{=}1024$ ($T{=}512$) on five models spanning four families. $\Delta\%$ is relative to each model's own uncompressed baseline. All cells are $3$-seed means except where noted.}
  \label{tab:freezone}
  \small
  \begin{tabular}{lrrrrr}
    \toprule
    Model & baseline & $2\times$ & $3\times$ & $4\times$ & $5\times$ \\
    \midrule
    Mistral-7B (GQA)
      & $6.28$
      & $6.28$\,{\scriptsize($-0.02\%$)}
      & $6.29$\,{\scriptsize($+0.04\%$)}
      & $6.64$\,{\scriptsize($+5.59\%$)}
      & $7.01$\,{\scriptsize($+11.56\%$)} \\
    LLaMA-2-13B (MHA)
      & $5.39$
      & $5.39$\,{\scriptsize($-0.02\%$)}
      & $5.39$\,{\scriptsize($-0.01\%$)}
      & $6.86$\,{\scriptsize($+27.28\%$)}
      & $9.07$\,{\scriptsize($+68.29\%$)} \\
    OLMoE-1B-7B$^{\dagger}$ (MoE)
      & $7.26$
      & $7.27$\,{\scriptsize($+0.02\%$)}
      & $7.27$\,{\scriptsize($+0.11\%$)}
      & $7.48$\,{\scriptsize($+3.00\%$)}
      & -- \\
    LLaMA-3.1-8B$^{\ddagger}$ (GQA)
      & $7.28$
      & $7.28$\,{\scriptsize($-0.01\%$)}
      & $7.29$\,{\scriptsize($+0.12\%$)}
      & $7.96$\,{\scriptsize($+9.33\%$)}
      & $11.02$\,{\scriptsize($+51.27\%$)} \\
    Qwen2.5-14B$^{*}$ (GQA)
      & $6.65$
      & $6.65$\,{\scriptsize($+0.02\%$)}
      & $6.66$\,{\scriptsize($+0.17\%$)}
      & $7.96$\,{\scriptsize($+19.63\%$)}
      & $9.55$\,{\scriptsize($+43.52\%$)} \\
    \bottomrule
  \end{tabular}
  \\[4pt]
  \footnotesize
  $^{\dagger}$\,WikiText-2 only, $100$ chunks, ratios $2$--$4$ only.
  $^{\ddagger}$\,FlashJoLT.
  $^{*}$\,FlashJoLT, single seed (seed~$0$), $100$ chunks.
\end{table}

\begin{table}[t]
  \centering
  \caption{Cross-method perplexity on Mistral-7B at $n{=}1024$ ($150$ chunks, $3$-seed mean). $\Delta\%$ is relative to the grid's uncompressed baseline of $6.76$. Quantizers operate at a single fixed ratio. This grid uses a different evaluation subset from Table~\ref{tab:freezone}. LLaMA-2 results in Appendix~\ref{app:recon}.}
  \label{tab:baselines}
  \small
  \begin{minipage}[t]{0.52\textwidth}
    \centering
    \textbf{Low-rank methods}\\[4pt]
    \begin{tabular}{lccc}
      \toprule
      Method & $2\times$ & $3\times$ & $4\times$ \\
      \midrule
      \textbf{JoLT} & $6.76$\,{\scriptsize($+0.01$)} & $6.76$\,{\scriptsize($+0.04$)} & $7.14$\,{\scriptsize($+5.65$)} \\
      \textbf{FlashJoLT} & $6.76$\,{\scriptsize($+0.00$)} & $6.79$\,{\scriptsize($+0.38$)} & $7.14$\,{\scriptsize($+5.59$)} \\
      xKV & $7.22$\,{\scriptsize($+6.75$)} & $7.65$\,{\scriptsize($+13.2$)} & $8.07$\,{\scriptsize($+19.4$)} \\
      Palu & $7.24$\,{\scriptsize($+7.06$)} & $7.61$\,{\scriptsize($+12.6$)} & $7.85$\,{\scriptsize($+16.1$)} \\
      ReCalKV & $7.25$\,{\scriptsize($+7.27$)} & $7.65$\,{\scriptsize($+13.1$)} & $7.90$\,{\scriptsize($+16.9$)} \\
      KQ-SVD & $7.12$\,{\scriptsize($+5.27$)} & $7.35$\,{\scriptsize($+8.67$)} & $7.60$\,{\scriptsize($+12.4$)} \\
      \bottomrule
    \end{tabular}
  \end{minipage}%
  \hfill
  \begin{minipage}[t]{0.46\textwidth}
    \centering
    \textbf{Quantizers}\\[4pt]
    \begin{tabular}{lcc}
      \toprule
      Method & ratio & PPL ($\Delta\%$) \\
      \midrule
      KIVI $2$-bit & $4.20\times$ & $6.88$\,{\scriptsize($+1.70$)} \\
      TurboQuant & $3.82\times$ & $6.78$\,{\scriptsize($+0.28$)} \\
      int4 p.c. & $3.97\times$ & $6.93$\,{\scriptsize($+2.42$)} \\
      int8 p.c. & $1.99\times$ & $6.76$\,{\scriptsize($-0.00$)} \\
      fp8 e4m3 & $2.00\times$ & $6.76$\,{\scriptsize($+0.01$)} \\
      \bottomrule
    \end{tabular}
  \end{minipage}
\end{table}

\subsection{Downstream evaluation}
\label{sec:downstream}
\begin{figure}[t]
  \centering
  \includegraphics[width=0.70\linewidth]{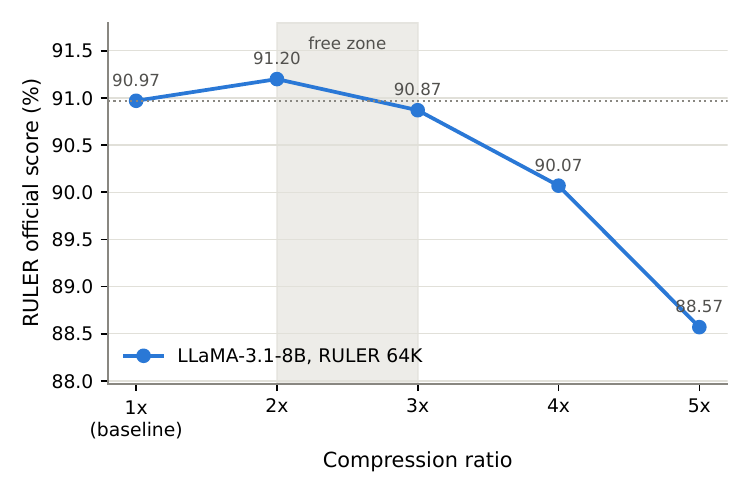}
  \caption{RULER score versus compression ratio on LLaMA-3.1-8B at $64$K context (FlashJoLT, $3$ dataset seeds, $600$ rows per arm). The $2$--$3\times$ free zone is shaded.}
  \label{fig:ruler_64k}
\end{figure}
\paragraph{Long-context retrieval (RULER, $64$K).}
On a $4$-task subset of RULER~\citep{hsieh2024ruler} ($50$ samples per task, recall-based accuracy with partial credit), FlashJoLT is near-lossless through $3\times$ at $64$K context on LLaMA-3.1-8B (Figure~\ref{fig:ruler_64k}).
At $2\times$ the compressed model scores $91.20$ against the $90.97$ uncompressed baseline ($\Delta = {+}0.23$~pp), within noise.
At $3\times$ the delta is $-0.10$~pp, also within noise.
Beyond the free zone, retrieval degrades smoothly: $4\times$ costs $-0.90$~pp and $5\times$ reaches $88.57$ ($\Delta = {-}2.40$~pp), with the full ladder monotone.%
\footnote{$95\%$ CIs from a seed-stratified percentile bootstrap ($N_\mathrm{boot}{=}200{,}000$ resamples $\times$ $5$ independent RNG streams). All RULER arms use $3$ dataset seeds and $600$ rows per arm.}

At $5\times$, the RULER score drops by only $2.40$~pp despite a $+51.27\%$ perplexity degradation at the same compression ratio (Table~\ref{tab:freezone}).
Perplexity scores every next token and is dominated by the hardest positions, while RULER scores a small number of retrieval-critical tokens.
The two metrics therefore measure different slices of model quality and need not degrade together.
Per-context-length results for Mistral and additional LLaMA-3.1 arms appear in Appendix~\ref{app:downstream}.

\paragraph{Reasoning (GSM8K).}
On GSM8K~\citep{cobbe2021gsm8k} with $8$-shot chain-of-thought prompting~\citep{wei2022cot} and exact-match scoring ($n{=}1319$ per cell), JoLT is near-lossless at $2\times$ and $3\times$ on both Mistral and LLaMA-2 (Table~\ref{tab:gsm8k_own} in Appendix~\ref{app:downstream}).
All free-zone deltas fall within their $95\%$ confidence intervals of zero, with the largest shift at $-1.21$~pp (Mistral $3\times$).
At $4\times$ accuracy drops sharply on Mistral ($-8.49$~pp), consistent with the perplexity cliff, while LLaMA-2 degrades more gently ($-2.20$~pp).
A matched-memory comparison against int4, KIVI, and TurboQuant appears in Appendix~\ref{app:downstream}.

\paragraph{Real-world long context (LongBench, $16$K).}
On six LongBench~\citep{bai2024longbench} tasks at $16$K context on Mistral-7B (multi-document QA, summarization, and few-shot learning, $200$ examples per task, seed $0$), FlashJoLT stays within $0.52$~pp of the uncompressed pooled score through $4\times$.
The pooled mean drifts by $-0.09$, $+0.05$, and $-0.52$~pp at $2\times$, $3\times$, and $4\times$, with no single task drifting by more than $0.91$~pp at any ratio (Appendix~\ref{app:downstream}, Table~\ref{tab:longbench}).
The evidence covers one model and one seed.

\subsection{Ablations}
\label{sec:ablations}

We ablate the design choices on Mistral-7B-v0.3.
Full grids across $n \in \{512, 1024, 2048\}$ are in Appendix~\ref{app:ablations} (Tables~\ref{tab:ablation_a1} and~\ref{tab:ablation_a5}).
Perplexities below are at $n{=}1024$ on the ablation evaluation subset unless stated.
\paragraph{Rotated residual.}
Dropping the residual entirely (rank-only compression) raises $2\times$ perplexity from $6.54$ to $7.27$, a gap that additional rank cannot close.
\paragraph{Allocation strategy.}
The joint Lagrangian allocation matches K/V-decoupled allocation at $2\times$ and trails it at $4\times$ ($6.93$ against $6.72$), where the budget is generous enough that rebalancing between keys and values has little effect.
At $8\times$ the joint solve pulls ahead ($8.14$ against $8.42$ at $n{=}1024$, $0.45$~PPL gap at $n{=}512$), where the tighter budget makes K/V rebalancing valuable.
At $n{=}2048$ the two stay within $0.17$~PPL of each other.
\paragraph{Per-group ranks.}
Per-group rank allocation beats uniform rank at all $15$ grid cells (ratios $2$, $3$, $4$, $6$, $8$ at three context lengths), with a peak advantage of $0.97$~PPL at $n{=}512$, $4\times$.
\paragraph{Residual bit-width.}
Residual bits saturate at four: the step from $4$ to $8$ bits improves perplexity by less than $0.001$ at $2\times$.

A further ablation on FlashJoLT versus JoLT parity appears in Appendix~\ref{app:fast}.

\subsection{Fused decode kernel}
\label{sec:efficiency}

Reconstructing dense keys and values from the compressed factors at each decode step would erase the memory savings.
The FlashJoLT decode kernel evaluates attention directly over the stored factors and packed residual, so no dense KV buffer is ever written to global memory.

The kernel follows the flash-decoding pattern: each program owns one KV head and a range of token blocks and carries one online-softmax state across them.
For each block, it rebuilds the key tile on chip from the token factor, core, and feature factor.
The residual is unpacked from its low-bit codes and derotated in the same pass, and RoPE is applied to the reconstructed tile before scoring against the query.
Values are handled differently: softmax weights are accumulated against the value token factor in rank space, only the value residual is decoded per block, and the rank-sized partial state is expanded through the core and feature factor once per head after the block states are merged.
Newly generated tokens live in a small dense tail that shares the same softmax state.

On the key side, the position-dependent RoPE does not commute with the token-mode factorization, so keys must be fully reconstructed per tile rather than scored in rank space as values are.
This is the main source of latency overhead and is discussed further in Section~\ref{sec:conclusion}.

\paragraph{Memory.}
On Mistral-7B-v0.3 at context length $8{,}192$, peak decode memory of the KV representation is $1.43\times$, $2.40\times$, $3.61\times$, and $4.73\times$ below the dense bf16 cache at the $2\times$, $3\times$, $5\times$, and $8\times$ compression targets.%
\footnote{The $8\times$ target overshoots to $9.8\times$ achieved ratio.}

\paragraph{Latency.}
The fused kernel is slower than uncompressed attention at every measured point.
In a microbenchmark of one Mistral-7B attention layer at batch $1$ on an A100 80GB, its latency is $7.5\times$ to $65.9\times$ the dense baseline (PyTorch scaled-dot-product attention over a resident bf16 cache) across contexts $512$ to $8{,}192$ at $2\times$, $3\times$, and $5\times$ targets.
Against a reconstruct-then-attend path, the fused kernel's latency ratio is $0.7\times$ to $2.2\times$, falling below $1$ only at context $8{,}192$.
End-to-end task latency confirms the overhead: on RULER NIAH with Mistral-7B at $8{,}192$ context (median over $250$ tasks, including compression and per-step reconstruction), the compressed path takes $20.7$s and $18.2$s at $2\times$ and $3\times$ against $1.6$s uncompressed.
At batch $1$ the resident dense cache is read faster than the factors can be contracted, so the kernel's current value is memory capacity, not speed.
We did not measure tokens-per-second throughput or batched serving, where the memory savings would translate into higher concurrency.

\section{Conclusion}
\label{sec:conclusion}

JoLT shows that low-rank compression is competitive with quantization in the near-lossless regime.
Across five models, partial Tucker decomposition paired with a rotated low-bit residual holds a $2$--$3\times$ free zone with less than $0.2\%$ perplexity drift, and a single Lagrangian dual makes this work at any target byte budget without per-model tuning.
The two approaches are complementary: quantizing the Tucker factors or the model weights can push compression further.

The fused decode kernel delivers the memory savings at serving time, reducing peak KV memory by up to $4.73\times$, but the latency cost is substantial.
The main bottleneck is RoPE: because the position-dependent rotation does not commute with the token-mode factorization, keys must be fully reconstructed per tile rather than scored in rank space as values are.
Folding the rotation into the factor scoring would remove this per-tile reconstruction and is the primary open problem on the kernel side.

Two further limitations point to future work.
The Lagrangian surrogate prices all Frobenius mass equally and does not model how key errors pass through the softmax, which leads to over-allocation of key rank on some models past the free zone.
A tighter surrogate or an improved cap policy would extend the near-lossless range beyond $3\times$.
The current pipeline compresses the prefill cache in one global pass, which suits prompt caching but does not extend to streaming autoregressive generation, where efficient online rank updates without full SVD recomputation remain an open problem.


\section*{Reproducibility Statement}

Source code for JoLT, FlashJoLT, and the baseline harnesses used in this paper is available at \url{https://github.com/rahulk98/JoLT-Master-Thesis}.
The method is fully specified in Section~\ref{sec:method} (layer grouping, the partial Tucker backbone, the rotated residual, and the allocator in Equations~\eqref{eq:bytes} to~\eqref{eq:lagrangian}), and the randomized variant and its cap policy are specified in Section~\ref{sec:fast} and Appendix~\ref{app:fast}.
Evaluation protocols, seeds, and evaluated-token counts are stated alongside each experiment (Sections~\ref{sec:results} and~\ref{sec:downstream}; Appendices~\ref{app:ppl}, \ref{app:downstream}, and~\ref{app:recon}); all five models are public checkpoints and all datasets (WikiText-2, C4, GSM8K, RULER, LongBench) are public.
GPU experiments ran on NVIDIA A100 GPUs (40\,GB PCIe, 40\,GB SXM4, and 80\,GB PCIe) and, for the $4$K and $8$K LLaMA-3.1-8B RULER rows of Table~\ref{tab:ruler_grid}, on one RTX 3090, under PyTorch~2.5.1 with CUDA~12.1, with models in bf16 and decomposition numerics in float32; the fused-kernel memory measurements reproduce bit-identically across the two A100 PCIe SKUs.

\section*{AI Use Statement}

The authors used generative AI tools during this work. LLM-based coding
assistants were used to assist with software implementation and experiment
scaffolding, generative AI tools were used to assist with manuscript
drafting and editing, and AI-assisted search was used to retrieve related
work and to check citation records against publisher metadata. Generative AI was not used to formulate the research
questions, design the proposed method or experimental protocol, analyze or
interpret experimental results, or formulate the paper's scientific claims. All
AI-assisted code and text were reviewed and verified by the authors. The authors
take full responsibility for the final content of the paper.

\clearpage
\bibliographystyle{unsrtnat}
\bibliography{references}

\newpage
\appendix
\raggedbottom
\section{Fast variant: speedup and cap calibration}
\label{app:fast}

\paragraph{Isolated compression-stage speedup.}
Table~\ref{tab:fast_speedup} reports the speedup of the randomized token-mode SVD in isolation on Mistral at $T{=}1024$ with $q_{\mathrm{cap}}{=}32$, measured on pre-grouped tensors with the allocator already swapped.
These are isolated stage timings, not end-to-end production figures.
LLaMA-2 gives $5.24\times$ to $9.97\times$ on the same stage.
On the same isolated path, the reconstruction-error gap between FlashJoLT and JoLT in the free zone at $T{=}1024$ is $|\Delta_K| \le 0.0093$ and $|\Delta_V| \le 0.0011$ on Mistral and $|\Delta_K| \le 0.0089$ and $|\Delta_V| \le 0.0031$ on LLaMA-2 at $2$--$3\times$.

Independent wall-clock timing of the two production compressors over the complete pipeline, including layer grouping, gives end-to-end speedups of $1.04\times$ (cache length $821$, $C{=}3.7$) and $3.97\times$ (cache length $1024$, $C{=}4$) on an A100 SXM4 40\,GB.
The token-mode SVD runs inside both the allocation and the fit stages, and FlashJoLT changes both.
No single-stage speedup therefore summarizes the end-to-end gain, and the isolated figures in Table~\ref{tab:fast_speedup} do not transfer to the production path.

\begin{table}[H]
  \centering
  \caption{Isolated speedup of the randomized token-mode SVD stage (Mistral, $T{=}1024$, $q_{\mathrm{cap}}{=}32$).}
  \label{tab:fast_speedup}
  \begin{tabular}{lrrrrrrrr}
    \toprule
    Target $C$ & $2$ & $3$ & $4$ & $5$ & $6$ & $7$ & $8$ & $10$ \\
    \midrule
    isolated speedup & $8.87\times$ & $9.30\times$ & $5.25\times$ & $7.76\times$ & $6.87\times$ & $13.57\times$ & $7.57\times$ & $12.86\times$ \\
    \bottomrule
  \end{tabular}
\end{table}

\paragraph{Cap calibration.}
The cap policy of Equation~\eqref{eq:qcap} is a no-op for $T \le 1024$ relative to the production default it replaced ($32$ for $C \le 4$, $64$ for $C > 4$), so every short-context number in this paper is unchanged by it.
The isolated sweep above fixes $q_{\mathrm{cap}}{=}32$ at every $C$ by design.
A calibration sweep on real Mistral KV confirms the policy is safe: for $C \le 5$ the reconstruction-error gate $\Delta_K \le 0.025$ passes at every context length from $512$ to $8192$.

\paragraph{Why the cap helps past the free zone.}
On LLaMA-2-13B at context length $n{=}1024$ and $4\times$, JoLT reaches $+27.28\%$ perplexity (Table~\ref{tab:freezone}) while FlashJoLT reaches $+14.16\%$ (Table~\ref{tab:ppl_baselines_llama}).
To separate the cap from the randomized SVD, we run three arms on the same $150$ chunks with three seeds each: FlashJoLT, JoLT, and JoLT with the token rank capped at $32$ and nothing else changed.
Against the matched uncompressed perplexity of $5.56$, the three arms land at $+14.21\%$, $+26.47\%$, and $+14.46\%$.
The capped exact solver matches FlashJoLT within $0.25$~pp and sits $12$~pp from the uncapped one, while reconstruction error is the same in all three ($0.12$ on keys, $0.08$ on values).

The difference is the allocation.
Without the cap, the allocator assigns token rank $55$ to $81$ on keys in six of the fourteen layer groups and drops their residual entirely, spending $18.8$~MB on key factors and $23.7$~MB on key residual.
Under the cap, the same groups spend $7.1$~MB on key factors and $34.8$~MB on key residual.
Value allocation is unchanged.
This is the surrogate limitation described in Section~\ref{sec:method}: on LLaMA-2 keys past the free zone, a key residual at $2$ to $4$ bits protects perplexity better than the extra rank the same bytes would buy.
Because RoPE preserves the Frobenius norm, the surrogate scores pre-RoPE and post-RoPE keys identically, but the rotated error can still matter through how it aligns with the queries.
An attention-logit-weighted surrogate would test this hypothesis, and we leave the mechanism open.
The randomized SVD adds only seed noise ($2.1$~pp spread for FlashJoLT against $0.02$~pp for the capped exact arm) around the same mean.

\section{Full perplexity grids}
\label{app:ppl}

This appendix reports the full perplexity grids of JoLT (exact SVD) across context lengths, referenced from Section~\ref{sec:results}.
All cells are corpus-level perplexity over the full $300$-chunk evaluation set.
$\Delta\%$ is relative to the uncompressed baseline at the same context length.
FlashJoLT matches these values within the near-lossless free zone (Appendix~\ref{app:fast}).

On Mistral (Table~\ref{tab:ppl_mistral}) the degradation past the free zone is gradual at every context length.
On LLaMA-2 (Table~\ref{tab:ppl_llama2}) the step between $4\times$ and $5\times$ is sharp, and perplexity peaks at $7\times$ at $n{=}512$ and $n{=}1024$ and at $8\times$ at $n{=}2048$, then falls through $10\times$.
An allocation log on $10$ chunks at $n{=}1024$ from $6\times$ to $10\times$ shows only monotone changes: no key cell stores residual bits at any of these targets and key ranks shrink, while value cells move from four residual bits to zero.
The allocation trend does not explain the non-monotone perplexity, which we attribute to a mismatch between the Frobenius surrogate and downstream perplexity (Appendix~\ref{app:fast}).

\paragraph{Cross-family degradation detail.}
Above the free zone, the degradation pattern is model-specific.
The corpus-level $8\times/3\times$ perplexity multiplier at $n{=}1024$ measures the steepness of high-ratio degradation.
Mistral is the mildest at $1.24\times$ and Qwen intermediate at $1.70\times$, while both LLaMA models are steep at $2.01\times$ (LLaMA-2) and $2.31\times$ (LLaMA-3.1).
Mistral and LLaMA-2 use JoLT (Tables~\ref{tab:ppl_mistral} and~\ref{tab:ppl_llama2}).
LLaMA-3.1 and Qwen use FlashJoLT, with the $8\times$ values given under Table~\ref{tab:freezone}.
This ordering does not track attention type or KV head count: LLaMA-3.1-8B, a GQA model with the same $8$~KV heads as Mistral, degrades as steeply as LLaMA-2-13B (MHA, $40$~KV heads), while Qwen (also GQA, $8$~KV heads) falls between the two.
Any operating point above $3\times$ therefore requires per-model validation rather than an architecture rule.

\paragraph{OLMoE (Mixture-of-Experts).}
OLMoE-1B-7B-0924~\citep{muennighoff2024olmoe} is a sparse MoE whose sparsity is FFN-only: the attention projections are standard dense layers, so the per-layer KV cache is a standard tensor and JoLT applies without modification.
Table~\ref{tab:olmoe_ppl} reports perplexity against the uncompressed baseline, both as $3$-seed means.
At $2\times$ the cost is $+0.02\%$ at both context lengths, and at $3\times$ it is $+0.11\%$ to $+0.14\%$.
At $4\times$ the drift is a graceful $+3.0\%$ to $+3.2\%$ with no cliff.
The seed spread is tight (per-seed $2\times$ deltas span $+0.005\%$ to $+0.045\%$), so near-losslessness is not a single-seed artifact.

\begin{table}[H]
  \centering
  \caption{JoLT perplexity on Mistral-7B-v0.3 (GQA) across context lengths; $\Delta\%$ relative to the uncompressed baseline.}
  \label{tab:ppl_mistral}
  \begin{tabular}{lrrr}
    \toprule
    Ratio & $n{=}512$ & $n{=}1024$ & $n{=}2048$ \\
    \midrule
    baseline & $6.96$ & $6.28$ & $6.01$ \\
    $2\times$ & $6.96$ ($+0.07\%$) & $6.28$ ($-0.02\%$) & $6.01$ ($+0.01\%$) \\
    $3\times$ & $6.97$ ($+0.11\%$) & $6.29$ ($+0.04\%$) & $6.01$ ($+0.07\%$) \\
    $4\times$ & $7.26$ ($+4.37\%$) & $6.64$ ($+5.59\%$) & $6.28$ ($+4.63\%$) \\
    $5\times$ & $7.78$ ($+11.81\%$) & $7.01$ ($+11.56\%$) & $6.53$ ($+8.76\%$) \\
    $8\times$ & $8.54$ ($+22.69\%$) & $7.78$ ($+23.77\%$) & $7.18$ ($+19.48\%$) \\
    $10\times$ & $8.93$ ($+28.36\%$) & $8.00$ ($+27.29\%$) & $7.46$ ($+24.17\%$) \\
    \bottomrule
  \end{tabular}
\end{table}

\begin{table}[H]
  \centering
  \caption{JoLT perplexity on LLaMA-2-13B (MHA) across context lengths; $\Delta\%$ relative to the uncompressed baseline.}
  \label{tab:ppl_llama2}
  \begin{tabular}{lrrr}
    \toprule
    Ratio & $n{=}512$ & $n{=}1024$ & $n{=}2048$ \\
    \midrule
    baseline & $5.95$ & $5.39$ & $5.19$ \\
    $2\times$ & $5.95$ ($+0.04\%$) & $5.39$ ($-0.02\%$) & $5.19$ ($+0.01\%$) \\
    $3\times$ & $5.96$ ($+0.08\%$) & $5.39$ ($-0.01\%$) & $5.19$ ($+0.05\%$) \\
    $4\times$ & $6.78$ ($+13.94\%$) & $6.86$ ($+27.28\%$) & $6.05$ ($+16.72\%$) \\
    $5\times$ & $10.76$ ($+80.70\%$) & $9.07$ ($+68.29\%$) & $7.43$ ($+43.26\%$) \\
    $7\times$ & $13.28$ ($+123.10\%$) & $11.04$ ($+104.93\%$) & $8.62$ ($+66.31\%$) \\
    $8\times$ & $12.72$ ($+113.64\%$) & $10.83$ ($+100.90\%$) & $8.72$ ($+68.21\%$) \\
    $10\times$ & $11.34$ ($+90.47\%$) & $9.04$ ($+67.82\%$) & $7.64$ ($+47.28\%$) \\
    \bottomrule
  \end{tabular}
\end{table}

\begin{table}[H]
  \centering
  \caption{JoLT perplexity on OLMoE-1B-7B-0924 (MoE) across context lengths, mean over $3$ seeds; WikiText-2 only, $100$ chunks per seed, a different chunk subset per seed; $\Delta\%$ relative to the uncompressed baseline.}
  \label{tab:olmoe_ppl}
  \begin{tabular}{lrr}
    \toprule
    Ratio & $n{=}512$ & $n{=}1024$ \\
    \midrule
    baseline & $7.9414$ & $7.2635$ \\
    $2\times$ & $7.9431$ ($+0.02\%$) & $7.2652$ ($+0.02\%$) \\
    $3\times$ & $7.9527$ ($+0.14\%$) & $7.2711$ ($+0.11\%$) \\
    $4\times$ & $8.1945$ ($+3.19\%$) & $7.4812$ ($+3.00\%$) \\
    \bottomrule
  \end{tabular}
\end{table}

\section{Downstream details}
\label{app:downstream}

This appendix collects the downstream task details referenced from Section~\ref{sec:downstream}.

\paragraph{GSM8K evaluation.}
Table~\ref{tab:gsm8k_own} reports JoLT's GSM8K accuracy on Mistral-7B and LLaMA-2-13B at its own compression ratios.
On both models, $2\times$ and $3\times$ are within noise of the uncompressed baseline, with confidence intervals that include zero.
At $4\times$, Mistral drops by $8.49$~pp, consistent with the perplexity cliff past the free zone.
LLaMA-2 at $4\times$ degrades more gently ($-2.20$~pp), with the confidence interval still touching zero.

\begin{table}[H]
  \centering
  \caption{GSM8K accuracy (\%), $8$-shot chain-of-thought, exact match, JoLT versus the uncompressed baseline ($n{=}1319$ per cell). Mistral $2\times$ and $3\times$ are identical across seeds $\{0,1,2\}$, all other cells are seed~$0$.}
  \label{tab:gsm8k_own}
  \begin{tabular}{lrrrr}
    \toprule
    Model & ratio & accuracy & $\Delta$ (pp) & $95$\% CI (pp) \\
    \midrule
    Mistral-7B (GQA) & baseline & $40.56$ & & \\
    & $2\times$ & $40.11$ & $-0.45$ & $[-1.52,\;{+}0.61]$ \\
    & $3\times$ & $39.35$ & $-1.21$ & $[-2.65,\;{+}0.15]$ \\
    & $4\times$ & $32.07$ & $-8.49$ & $[-11.1,\;{-}5.9]$ \\
    \midrule
    LLaMA-2-13B (MHA) & baseline & $28.13$ & & \\
    & $2\times$ & $28.43$ & ${+}0.30$ & $[-0.8,\;{+}1.4]$ \\
    & $3\times$ & $28.51$ & ${+}0.38$ & $[-0.9,\;{+}1.7]$ \\
    & $4\times$ & $25.93$ & $-2.20$ & $[-4.40,\;{+}0.08]$ \\
    \bottomrule
  \end{tabular}
\end{table}

\paragraph{GSM8K matched-memory comparison.}
Table~\ref{tab:gsm8k_matched} compares JoLT against fixed-rate quantizers at the byte rate of int4 per-channel quantization, roughly $4\times$, on Mistral-7B.
At this rate the quantizers are competitive: KIVI 2-bit scores $40.18\%$ ($+3.11$~pp over int4's $37.07\%$) at a slightly looser budget, and TurboQuant scores $38.89\%$ ($+1.82$~pp).
JoLT at the same budget achieves $35.94\%$ at an achieved ratio of $4.1\times$ ($-1.14$~pp versus int4).
The int4-native byte rate forces JoLT past its free zone at short context.
At $64$K and above, where the KV cache is the actual bottleneck, FlashJoLT costs under one point at $4\times$ (Section~\ref{sec:downstream}).

\begin{table}[H]
  \centering
  \caption{GSM8K accuracy on Mistral-7B at matched memory ($n{=}1319$). Top: JoLT at its own ratios, $\Delta$ versus the uncompressed baseline. Bottom: methods near the int4 per-channel byte rate, $\Delta$ versus int4. B/token is serialized prefill-KV bytes per token. KIVI and TurboQuant sit slightly above int4's bytes. Deltas are computed from unrounded accuracies.}
  \label{tab:gsm8k_matched}
  \begin{tabular}{lrrrrl}
    \toprule
    Method & B/token & ratio & acc (\%) & $\Delta$ (pp) & \\
    \midrule
    \multicolumn{6}{l}{\emph{JoLT own ratio ($\Delta$ vs uncompressed)}} \\
    uncompressed    & $131072.0$ & $1.0\times$ & $40.56$ &         & \\
    JoLT $2\times$  &  $65540.4$ & $2.0\times$ & $40.11$ & $-0.45$ & \\
    JoLT $3\times$  &  $43673.4$ & $3.0\times$ & $39.35$ & $-1.21$ & \\
    \midrule
    \multicolumn{6}{l}{\emph{Near int4 bytes ($\Delta$ vs int4)}} \\
    KIVI $2$-bit       & $35975.6$ & $3.6\times$ & $40.18$ & ${+}3.11$ & $\star$ \\
    TurboQuant       & $34304.0$ & $3.8\times$ & $38.89$ & ${+}1.82$ &  \\
    int4 per-ch.\    & $33089.6$ & $4.0\times$ & $37.07$ &    ref    & ref \\
    JoLT & $32066.5$ & $4.1\times$ & $35.94$ & $-1.14$ &  \\
    \bottomrule
  \end{tabular}
  \\[4pt]
  \footnotesize
  $\star$~KIVI sits at a looser budget.
\end{table}

\paragraph{RULER full grid.}
Table~\ref{tab:ruler_grid} reports the full RULER grid across context lengths and compression ratios.
On Mistral-7B at $4$K and $8$K ($3$ dataset seeds, $1200$ rows per arm), the $2\times$ and $3\times$ arms track the baseline closely: at $8$K the partial-credit deltas are $-0.20$~pp and $-0.70$~pp on a $97.37$ baseline.
Extending to $16$K and $32$K ($2$ dataset seeds, $800$ rows per arm), both arms remain near-lossless, with pooled deltas of $-0.45$~pp at $2\times$ and $-0.225$~pp at $3\times$.
A single-seed $4\times$ stress arm on Mistral shows the penalty shrinking with context: $-6.10$~pp at $4$K, $-0.60$~pp at $8$K, $-0.30$~pp at $16$K, and $-0.20$~pp at $32$K.

\begin{table}[H]
  \centering
  \caption{RULER partial-credit accuracy on the $4$-task subset~\citep{hsieh2024ruler}, FlashJoLT. LLaMA-3.1-8B at $64$K: $3$ dataset seeds, $600$ rows per arm. Mistral-7B: $3$ seeds at $4$K and $8$K, $2$ seeds at $16$K and $32$K. Mistral $4\times$ is a single-seed stress arm (seed~$42$), reported against its own seed's baseline. No $5\times$ arm was run on Mistral. LLaMA-3.1-8B at $4$K and $8$K was run at $5\times$ only, on an RTX 3090.}
  \label{tab:ruler_grid}
  \begin{tabular}{llrrrrr}
    \toprule
    Model & context & baseline & $2\times$ & $3\times$ & $4\times$ & $5\times$ \\
    \midrule
    \multicolumn{7}{l}{\emph{LLaMA-3.1-8B (GQA), $64$K, $3$ dataset seeds}} \\
    & $65536$
      & $90.97$
      & $91.20$\,{\scriptsize($+0.23$)}
      & $90.87$\,{\scriptsize($-0.10$)}
      & $90.07$\,{\scriptsize($-0.90$)}
      & $88.57$\,{\scriptsize($-2.40$)} \\[2pt]
    \midrule
    \multicolumn{7}{l}{\emph{Mistral-7B (GQA), $4$K to $32$K}} \\
    & $4096$  & $98.80$ & $98.80$\,{\scriptsize($+0.00$)} & $98.80$\,{\scriptsize($+0.00$)} & $92.70$\,{\scriptsize($-6.10$)} & -- \\
    & $8192$  & $97.37$ & $97.17$\,{\scriptsize($-0.20$)} & $96.67$\,{\scriptsize($-0.70$)} & $96.77$\,{\scriptsize($-0.60$)} & -- \\
    & $16384$ & $95.80$ & $95.60$\,{\scriptsize($-0.20$)} & $95.55$\,{\scriptsize($-0.25$)} & $95.50$\,{\scriptsize($-0.30$)} & -- \\
    & $32768$ & $85.90$ & $85.20$\,{\scriptsize($-0.70$)} & $85.70$\,{\scriptsize($-0.20$)} & $85.70$\,{\scriptsize($-0.20$)} & -- \\
    \midrule
    \multicolumn{7}{l}{\emph{LLaMA-3.1-8B (GQA), $4$K and $8$K, $3$ dataset seeds}} \\
    & $4096$  & $99.67$ & -- & -- & -- & $92.83$\,{\scriptsize($-6.83$)} \\
    & $8192$  & $99.67$ & -- & -- & -- & $98.53$\,{\scriptsize($-1.13$)} \\
    \bottomrule
  \end{tabular}
\end{table}

\paragraph{LongBench $16$K on Mistral-7B.}
Table~\ref{tab:longbench} reports per-task results for FlashJoLT versus the uncompressed baseline on six LongBench tasks at $16$K context ($200$ examples per task, seed $0$, official LongBench metrics).

\begin{table}[H]
  \centering
  \caption{LongBench $16$K, Mistral-7B, FlashJoLT versus baseline, per-task means with $\Delta$ against the uncompressed baseline in parentheses. Metrics: F1 for HotpotQA and 2WikiMQA, ROUGE-L for GovReport, QMSum and SAMSum, classification accuracy for TREC.}
  \label{tab:longbench}
  \begin{tabular}{lrrrr}
    \toprule
    task & baseline & $2\times$ & $3\times$ & $4\times$ \\
    \midrule
    hotpotqa (F1)      & $12.83$ & $12.72$ ($-0.11$) & $12.94$ ($+0.10$) & $12.21$ ($-0.63$) \\
    2wikimqa (F1)      & $11.77$ & $11.88$ ($+0.11$) & $11.62$ ($-0.15$) & $11.62$ ($-0.15$) \\
    gov\_report (RG-L) & $28.31$ & $28.24$ ($-0.07$) & $29.08$ ($+0.77$) & $27.40$ ($-0.91$) \\
    qmsum (RG-L)       & $22.88$ & $22.90$ ($+0.02$) & $22.97$ ($+0.09$) & $22.56$ ($-0.31$) \\
    samsum (RG-L)      & $39.87$ & $39.41$ ($-0.46$) & $39.36$ ($-0.51$) & $39.24$ ($-0.63$) \\
    trec (acc)         & $75.50$ & $75.50$ ($+0.00$) & $75.50$ ($+0.00$) & $75.00$ ($-0.50$) \\
    \midrule
    pooled mean        & $31.86$ & $31.77$ ($-0.09$) & $31.91$ ($+0.05$) & $31.34$ ($-0.52$) \\
    \bottomrule
  \end{tabular}
\end{table}
\section{Full ablation grids}
\label{app:ablations}

This appendix reports the context-resolved ablation grids referenced from Section~\ref{sec:ablations}.
Ablation perplexities are evaluated on a $100$-chunk subset ($51{,}100$ evaluated tokens per cell) and are comparable within this appendix but not with the full-corpus values of Table~\ref{tab:freezone}.

Table~\ref{tab:ablation} collects the headline cells at $n{=}1024$ (mean PPL over $3$ seeds).
Tables~\ref{tab:ablation_a1} and~\ref{tab:ablation_a5} resolve the allocation-strategy and per-group-rank ablations by context length.
The joint column of Table~\ref{tab:ablation_a1} and the per-group column of Table~\ref{tab:ablation_a5} are separate runs of the same configuration and agree to within $0.001$~PPL ($8.145$ against $8.144$ at $n{=}1024$, $8\times$).

In the allocation-strategy grid (Table~\ref{tab:ablation_a1}), the joint and K/V-decoupled solvers track each other closely at $2\times$ across all three context lengths.
The joint solver pulls ahead at $8\times$ where the tighter budget makes K/V rebalancing valuable, except at $n{=}2048$ where the two stay within $0.17$~PPL.
Per-group rank allocation (Table~\ref{tab:ablation_a5}) beats uniform rank at every cell, with the largest gap at short context and high compression ($0.97$~PPL at $n{=}512$, $4\times$).

\begin{table}[H]
  \centering
  \caption{Ablations on Mistral-7B-v0.3 at $n{=}1024$, mean PPL over $3$ seeds. Allocation strategy: joint, K/V-decoupled, and rank-only (no residual). Per-group versus uniform rank. Residual bits at $2\times$.}
  \label{tab:ablation}
  \begin{tabular}{lrrr}
    \toprule
    & $2\times$ & $4\times$ & $8\times$ \\
    \midrule
    \multicolumn{4}{l}{\emph{Allocation strategy}} \\
    \quad Joint (ours) & $6.54$ & $6.93$ & $8.14$ \\
    \quad K/V-decoupled & $6.54$ & $6.72$ & $8.42$ \\
    \quad Rank-only (no residual) & $7.27$ & $7.95$ & $8.65$ \\
    \midrule
    \multicolumn{4}{l}{\emph{Per-group vs.\ uniform rank}} \\
    \quad Per-group (ours) & $6.54$ & $6.93$ & $8.14$ \\
    \quad Uniform rank & $6.70$ & $7.47$ & $8.68$ \\
    \midrule
    \multicolumn{4}{l}{\emph{Residual bits at $2\times$ ($0$ / $2$ / $4$ / $8$ bit)}} \\
    \quad PPL & \multicolumn{3}{l}{$7.27 \;/\; 6.63 \;/\; 6.537 \;/\; 6.536$} \\
    \bottomrule
  \end{tabular}
\end{table}

\begin{table}[H]
  \centering
  \caption{Allocation strategy by context length (joint / K/V-decoupled / rank-only), mean PPL over $3$ seeds.}
  \label{tab:ablation_a1}
  \begin{tabular}{lrrr}
    \toprule
    $n$, ratio & joint & K/V-decoupled & rank-only \\
    \midrule
    $512$, $2\times$  & $6.532$ & $6.534$ & $7.412$ \\
    $512$, $4\times$  & $6.777$ & $6.699$ & $8.121$ \\
    $512$, $8\times$  & $7.917$ & $8.362$ & $8.463$ \\
    $1024$, $2\times$ & $6.536$ & $6.535$ & $7.269$ \\
    $1024$, $4\times$ & $6.928$ & $6.717$ & $7.946$ \\
    $1024$, $8\times$ & $8.145$ & $8.418$ & $8.649$ \\
    $2048$, $2\times$ & $5.981$ & $5.981$ & $6.394$ \\
    $2048$, $4\times$ & $6.259$ & $6.091$ & $6.855$ \\
    $2048$, $8\times$ & $7.102$ & $7.057$ & $7.388$ \\
    \bottomrule
  \end{tabular}
\end{table}

\begin{table}[H]
  \centering
  \caption{Per-group versus uniform rank by context length, mean PPL over $3$ seeds.}
  \label{tab:ablation_a5}
  \begin{tabular}{lrr}
    \toprule
    $n$, ratio & per-group & uniform \\
    \midrule
    $512$, $2\times$  & $6.532$ & $6.843$ \\
    $512$, $4\times$  & $6.777$ & $7.749$ \\
    $512$, $8\times$  & $7.917$ & $8.712$ \\
    $1024$, $2\times$ & $6.536$ & $6.702$ \\
    $1024$, $4\times$ & $6.927$ & $7.470$ \\
    $1024$, $8\times$ & $8.144$ & $8.680$ \\
    $2048$, $2\times$ & $5.981$ & $6.059$ \\
    $2048$, $4\times$ & $6.260$ & $6.509$ \\
    $2048$, $8\times$ & $7.102$ & $7.185$ \\
    \bottomrule
  \end{tabular}
\end{table}

\section{Cross-method reconstruction and perplexity benchmark}
\label{app:recon}

\paragraph{Baseline implementations.}
KIVI and xKV run the authors' released code, and ReCalKV runs the authors' head-reordering code.
Palu, KQ-SVD, and TurboQuant are our reimplementations from the papers, applied to the prompt's cache under the deviations listed below, so their numbers are not reproductions of the originally reported results.

The Palu arm applies Palu's grouped low-rank decomposition to the prompt's cache (a per-prompt SVD of the grouped cache rather than an offline factorization of the projection weights) with uniform per-group rank, in place of Palu's offline Fisher-information rank search.
The ReCalKV arm keeps ReCalKV's head reordering and grouped SVD for keys but applies a per-prompt SVD to the value cache in place of offline value calibration, with uniform rank in place of Fisher-based allocation.
The KQ-SVD arm computes its key factorization from the prompt's own queries rather than offline calibration queries and uses per-head SVD for values.
The TurboQuant arm uses uniform $4$ bits with no outlier-channel split, the inner-product variant for keys and the MSE variant for values.
Palu is evaluated as an fp16 low-rank method without optional latent quantization.
Quantizer key errors are measured on post-RoPE keys and low-rank key errors on pre-RoPE keys.

\paragraph{Reconstruction error.}
Tables~\ref{tab:recon_mistral} and~\ref{tab:recon_llama} report mean relative Frobenius reconstruction error at $T{=}1024$ for all methods on each model at $2\times$ and $4\times$.

At $2\times$, JoLT and FlashJoLT rank first and second among the low-rank methods on keys and values on both architectures.
The fixed-rate int8 baseline has lower key error ($0.0047$ on Mistral) but higher value error.
On Mistral values, JoLT reaches $0.00571$, roughly $27\times$ below Tucker without residual at $0.15546$, the best non-JoLT low-rank result.
The residual rotation is fixed across seeds, so every method varies across seeds only through the chunk draw.
The standard deviations shown for JoLT and FlashJoLT are at most $0.0003$, against a gap of at least $0.088$ to the nearest baseline.

At $4\times$, keys show a crossover: int4 per-channel outperforms JoLT on Mistral ($0.080$ against $0.118$) and LLaMA-2 ($0.077$ against $0.120$), while JoLT remains better on values ($0.090$ against $0.132$ and $0.087$ against $0.124$).
This does not affect the free-zone claim because the evaluated low-bit fixed-rate quantizers cannot operate in the $2$--$3\times$ band, but at $4\times$ the best method depends on whether key or value error dominates task quality.
Per-head SVD and KQ-SVD have bit-identical value errors because they share a value path.
Mistral int4 results use the source's three-decimal precision.

\begin{table}[H]
  \centering
  \caption{Relative Frobenius reconstruction error on Mistral-7B-v0.3 (GQA) at $T{=}1024$, mean over $3$ seeds; lower is better. Parentheses: across-seed standard deviation at $2\times$ for JoLT and FlashJoLT; the seed changes the reconstruction chunks drawn, not the residual rotation. int4 per-channel has no $2\times$ operating point.}
  \label{tab:recon_mistral}
  \begin{tabular}{lcccc}
    \toprule
    & \multicolumn{2}{c}{$2\times$} & \multicolumn{2}{c}{$4\times$} \\
    \cmidrule(lr){2-3} \cmidrule(lr){4-5}
    Method & K & V & K & V \\
    \midrule
    \textbf{JoLT (ours)} & $0.00903$\,{\scriptsize($.0002$)} & \textbf{$0.00571$}\,{\scriptsize($.0001$)} & $0.11844$ & \textbf{$0.08964$} \\
    \textbf{FlashJoLT (ours)} & \textbf{$0.00782$}\,{\scriptsize($.0003$)} & $0.00584$\,{\scriptsize($.0001$)} & \textbf{$0.11765$} & $0.09307$ \\
    \midrule
    xKV~\citep{chang2025xkv} & $0.10185$ & $0.20863$ & $0.16212$ & $0.37706$ \\
    Palu~\citep{chang2024palu} & $0.09912$ & $0.31842$ & $0.17219$ & $0.49365$ \\
    ReCalKV~\citep{yan2025recalkv} & $0.09731$ & $0.27728$ & $0.16940$ & $0.44446$ \\
    Tucker-only (ours, no residual) & $0.14027$ & $0.15546$ & $0.20661$ & $0.32525$ \\
    Per-head SVD & $0.13850$ & $0.42096$ & $0.24208$ & $0.60426$ \\
    KQ-SVD~\citep{lesens2025kqsvd} & $0.15496$ & $0.42096$ & $0.26287$ & $0.60426$ \\
    \midrule
    \multicolumn{5}{l}{\emph{Fixed-rate quantizer at native $\approx 3.97\times$}} \\
    int4 per-channel & n/a & n/a & $0.080$ & $0.132$ \\
    \bottomrule
  \end{tabular}
\end{table}

\begin{table}[H]
  \centering
  \caption{Relative Frobenius reconstruction error on LLaMA-2-13B (MHA) at $T{=}1024$. FlashJoLT and xKV are $3$-seed means, the other low-rank baselines single-seed. JoLT and Tucker-only were not run on this grid. Quantizers at their single evaluated setting.}
  \label{tab:recon_llama}
  \begin{tabular}{lcccc}
    \toprule
    & \multicolumn{2}{c}{$2\times$} & \multicolumn{2}{c}{$4\times$} \\
    \cmidrule(lr){2-3} \cmidrule(lr){4-5}
    Method & K & V & K & V \\
    \midrule
    \textbf{FlashJoLT (ours)} & \textbf{$0.007452$} & \textbf{$0.005614$} & \textbf{$0.120333$} & \textbf{$0.086566$} \\
    \midrule
    xKV~\citep{chang2025xkv} & $0.114161$ & $0.190025$ & $0.177292$ & $0.363146$ \\
    Palu~\citep{chang2024palu} & $0.116724$ & $0.318463$ & $0.189522$ & $0.478121$ \\
    ReCalKV~\citep{yan2025recalkv} & $0.114969$ & $0.224396$ & $0.186213$ & $0.380975$ \\
    Per-head SVD & $0.148635$ & $0.402574$ & $0.241318$ & $0.574452$ \\
    KQ-SVD~\citep{lesens2025kqsvd} & $0.166458$ & $0.402574$ & $0.263680$ & $0.574452$ \\
    \midrule
    \multicolumn{5}{l}{\emph{Fixed-rate quantizers at their evaluated setting}} \\
    int4 per-channel ($\approx 3.97\times$) & n/a & n/a & $0.077078$ & $0.123518$ \\
    KIVI $2$-bit~\citep{liu2024kivi} ($4.20\times$) & n/a & n/a & $0.200598$ & $0.383170$ \\
    TurboQuant~\citep{turboquant2025} ($3.82\times$) & n/a & n/a & $0.227269$ & $0.096594$ \\
    \bottomrule
  \end{tabular}
\end{table}

\paragraph{Perplexity on the same grid.}
Table~\ref{tab:baselines} in the main text reports the cross-method perplexity grid on Mistral.
Two baselines that do not appear in that table are Tucker-only (ours, no residual), which reaches $+10.81\%$ at $2\times$ and $+21.23\%$ at $4\times$, and Per-head SVD, which reaches $+8.39\%$ and $+15.74\%$ at the same ratios.

Table~\ref{tab:ppl_baselines_llama} reports the corresponding grid on LLaMA-2-13B.
All LLaMA-2 methods except Palu, ReCalKV, Per-head SVD, and KQ-SVD are $3$-seed means.
JoLT and Tucker-only were not run on this grid, so FlashJoLT represents the JoLT family.
Its $+14.16\%$ at $4\times$, rather than JoLT's $+27.28\%$ in Table~\ref{tab:freezone}, reflects the token-rank cap of Section~\ref{sec:fast}, which binds on LLaMA-2 beyond the free zone (Appendix~\ref{app:fast}).
The pattern matches Mistral: JoLT's advantage lies in the $2$--$3\times$ band where no fixed-rate quantizer operates, and near $4\times$ TurboQuant and KIVI outperform it on perplexity.

At $2\times$, fixed-rate $8$-bit baselines match free-zone quality on both models.
int8 per-channel achieves relative Frobenius error $0.0047$/$0.0077$ (K/V) on Mistral at $T{=}1024$ ($1.99\times$ achieved) and $0.0045$/$0.0073$ on LLaMA-2 ($1.99\times$), with perplexity within $0.02\%$ of the uncompressed baseline on both models.
fp8 e4m3 sits at exactly $2.00\times$ with reconstruction error roughly $6\times$ higher than int8 on keys and $3.5\times$ higher on values, but perplexity remains within $0.04\%$.
int8 per-channel stores one scale per row, so its ratio is $2T/(T{+}4) < 2$.
Neither format has an operating point above $2\times$.

\begin{table}[H]
  \centering
  \caption{Cross-method perplexity on LLaMA-2-13B (MHA) at $n{=}1024$ (WikiText-2 and C4, $150$ chunks, mean over seeds). Parentheses give the $\Delta\%$ against the grid's uncompressed perplexity of $5.56$. JoLT and Tucker-only were not run on this grid. Quantizers run only at their single evaluated setting, shown next to the name.}
  \label{tab:ppl_baselines_llama}
  \small
  \begin{tabular}{lccc}
    \toprule
    Method & $2\times$ & $3\times$ & $4\times$ \\
    \midrule
    \textbf{FlashJoLT (ours)} & $5.56$\,{\scriptsize($+0.01$)} & $5.56$\,{\scriptsize($+0.03$)} & $6.35$\,{\scriptsize($+14.16$)} \\
    \midrule
    xKV~\citep{chang2025xkv} & $8.74$\,{\scriptsize($+57.21$)} & $7.79$\,{\scriptsize($+40.14$)} & $7.71$\,{\scriptsize($+38.66$)} \\
    Palu~\citep{chang2024palu} & $6.28$\,{\scriptsize($+12.96$)} & $6.55$\,{\scriptsize($+17.76$)} & $6.75$\,{\scriptsize($+21.48$)} \\
    ReCalKV~\citep{yan2025recalkv} & $6.81$\,{\scriptsize($+22.59$)} & $7.02$\,{\scriptsize($+26.35$)} & $7.21$\,{\scriptsize($+29.71$)} \\
    Per-head SVD & $6.01$\,{\scriptsize($+8.04$)} & $6.24$\,{\scriptsize($+12.22$)} & $6.45$\,{\scriptsize($+16.12$)} \\
    KQ-SVD~\citep{lesens2025kqsvd} & $5.87$\,{\scriptsize($+5.52$)} & $6.04$\,{\scriptsize($+8.67$)} & $6.22$\,{\scriptsize($+11.99$)} \\
    \midrule
    \multicolumn{4}{l}{\emph{Fixed-rate quantizers at their evaluated setting (single operating point, shown in the $4\times$ column)}} \\
    KIVI $2$-bit~\citep{liu2024kivi} ($\approx 4.20\times$) & -- & -- & $5.65$\,{\scriptsize($+1.59$)} \\
    TurboQuant~\citep{turboquant2025} ($\approx 3.82\times$) & -- & -- & $5.57$\,{\scriptsize($+0.25$)} \\
    int4 per-channel ($\approx 3.97\times$) & -- & -- & $6.01$\,{\scriptsize($+8.08$)} \\
    \midrule
    \multicolumn{4}{l}{\emph{Fixed-rate $8$-bit at native rate (shown in the $2\times$ column)}} \\
    int8 per-channel ($\approx 1.99\times$) & $5.56$\,{\scriptsize($+0.02$)} & -- & -- \\
    fp8 e4m3 ($2.00\times$) & $5.56$\,{\scriptsize($+0.04$)} & -- & -- \\
    \bottomrule
  \end{tabular}
\end{table}

\end{document}